%% file: main.tex
\documentclass[11pt,letterpaper]{article}
\usepackage[T1]{fontenc}
\usepackage[margin=1.02in]{geometry}
\usepackage{amsmath,amsthm}
\usepackage{newtxtext,newtxmath}
\usepackage{courier}
\usepackage{microtype,graphicx,booktabs}
\usepackage[round,authoryear]{natbib}
\usepackage{caption}
\usepackage[hyphens]{url}
\usepackage[hidelinks,pdfauthor={Enliang Hu},pdftitle={When Is a Low-Rank Metric Update Globally Optimal?}]{hyperref}
\usepackage{bookmark}
\newcommand{\doi}[1]{doi: \href{https://doi.org/#1}{\begingroup\urlstyle{rm}\nolinkurl{#1}\endgroup}}
\newtheorem{theorem}{Theorem}
\newtheorem{proposition}[theorem]{Proposition}
\newcommand{\tr}{\operatorname{tr}}
\newcommand{\rank}{\operatorname{rank}}
\newcommand{\Kr}{\mathcal K_r}
\newcommand{\sym}{\operatorname{sym}}
\title{When Is a Low-Rank Metric Update Globally Optimal?}
\author{Enliang Hu}
\date{}
\begin{document}
\maketitle
\begin{abstract}
Can a learned distance metric be improved while keeping the same rank budget? We study positive semidefinite metric updates under a LogDet objective with convex supervision. Linearizing the supervised term gives an upper bound on suboptimality at the prescribed rank. At a stationary factor of full column rank with positive semidefinite Hessian, the bound vanishes when the eigengap at the rank cutoff exceeds the product of the supervision curvature bound and the smallest retained eigenvalue of the update. In a two-dimensional example, equality permits a nonglobal second-order stationary point and stronger supervision stabilizes a suboptimal local minimum. Under uniform spectral separation and an explicit contraction condition, a spectral iteration converges from any feasible start to the unique global minimizer. A fixed penalty also gives a correspondence between second-order stationary points of symmetric and asymmetric factorizations. On 180 supervised training problems, interval arithmetic places feasible metrics reconstructed from 174 split-solver outputs within $7\times10^{-13}$ of the fixed-rank optimum. The unrestricted numerical bound meets a tolerance of $10^{-7}$ in 51 cases. In one of the six remaining settings, a verified rank-two competitor lowers the objective by more than $1.23\times10^{-4}$.
\end{abstract}

\section{Introduction}
A low-rank correction to a distance metric is useful when the available measurements support only a small number of directions. Consider a metric $I+UU^\top$ on feature vectors in $\mathbb R^d$, with $U$ having $r$ columns. A proposed additional direction can lower the training loss even when these $r$ columns already give the best possible update. To decide whether to stop at width $r$, the relevant comparison is with the best direction that would have to be removed.

A simple example illustrates the issue. Minimize $\langle C,W\rangle-\log\det(I+W)$ over positive semidefinite matrices $W$ of rank at most one, with $C=\operatorname{diag}(1/4,1/2,3)$. The unique minimizer is $W=\operatorname{diag}(3,0,0)$. The gradient there is $\operatorname{diag}(0,-1/2,2)$. Its negative entry signals a genuine improvement at rank two. A test that requires a positive semidefinite gradient cannot certify the rank-one answer. The eigenvalues of $C$ settle this particular example. With supervision from relative distances, the favorable directions depend on the candidate itself.

We consider a LogDet objective with an additional convex supervised loss. Linearizing only the supervised term leaves a rank-constrained problem with a closed-form solution. This specialization of a composite primal--dual gap provides a certificate at any feasible iterate. Our main result determines when the certificate vanishes at a second-order stationary point. The proof rotates the weakest retained mode into an omitted direction and compares the resulting spectral improvement with the directional curvature of supervision. A sharp counterexample shows how an increase in that curvature creates a strict suboptimal local minimum. When spectral separation holds throughout the feasible set and supervision varies sufficiently slowly, the same spectral calculation also yields an iteration with a global convergence guarantee.

For asymmetric factors, a fixed penalty can force agreement at stationarity. Once the factors agree, their common range still has to be tested against other subspaces of the same dimension. On a handwritten-digit training problem, two fitted factors differ by only $1.24\times10^{-8}$ in Frobenius norm, yet a second rank-two metric attains a strictly lower objective. The same certificate can be applied to either metric and to the output of a projected solver.

\input{related}

\section{An Optimality Bound at the Prescribed Rank}
Write $\Kr=\{W\in\mathbb S^d: W\succeq0,\ \rank W\le r\}$ with $1\le r<d$. For symmetric matrices, $\langle A,B\rangle=\tr(AB)$. We order eigenvalues of an effective matrix increasingly, write $\|\cdot\|_2$ for the spectral norm and use $[x]_+=\max\{x,0\}$. The objective is
\begin{equation}
 J(W)=\langle C,W\rangle-\log\det(I+W)+h(W),\qquad W\in\Kr ,
 \label{eq:model}
\end{equation}
where $C\succ0$ and $h$ is convex and differentiable on the symmetric matrices under consideration. We write $J_r^\star=\inf_{V\in\Kr}J(V)$ and assume it is finite. For nonnegative supervision the objective is coercive on the closed set $\Kr$, so the infimum is attained. The linear and LogDet terms together penalize departure from a reference metric. In particular, with $C=I$ it is the LogDet divergence of $I+W$ from $I$, up to the constant already removed in~\eqref{eq:model}.

At a candidate $W$, set $K_W=C+\nabla h(W)$ and define
\begin{align}
 q_K(V)&=\langle K,V\rangle-\log\det(I+V),\nonumber\\
 s_r(K)&=\inf_{V\in\Kr}q_K(V),\quad
 \Delta_r(W)=q_{K_W}(W)-s_r(K_W). \label{eq:gap}
\end{align}
For $K\succ0$, let $k_1\le\cdots\le k_d$ and $v_i$ be its eigenpairs. The scalar cost $kt-\log(1+t)$ on $t\ge0$ is minimized by $t=(1/k-1)_+$. Hence
\begin{align}
 b(k)&=\begin{cases}1-k+\log k,&0<k<1,\\0,&k\ge1,\end{cases}\nonumber\\
 s_r(K)&=\sum_{i=1}^{r}b(k_i),\quad
 V_K=\sum_{i=1}^{r}(1/k_i-1)_+v_iv_i^\top. \label{eq:oracle}
\end{align}
Indeed, the trace term pairs the largest eigenvalues of $V$ with the smallest of $K$, and the minimum scalar cost increases with $k$ for $k<1$. At a repeated cutoff any minimizing subspace gives the same value. If $k_1\le0$, moving along its eigenvector makes $s_r(K)=-\infty$.

\begin{proposition}[Fixed-rank objective gap]\label{prop:gap}
For any $W\in\Kr$, $0\le J(W)-J_r^\star\le\Delta_r(W)$. If $K_W\succ0$ and $V=V_{K_W}$, one also has
\[
 [\Delta_r(W)-D_h(V,W)]_+\ \le\ J(W)-J_r^\star ,
\]
where $D_h(V,W)=h(V)-h(W)-\langle\nabla h(W),V-W\rangle$.
\end{proposition}

Convexity gives $J(V)\ge q_{K_W}(V)+h(W)-\langle\nabla h(W),W\rangle$ for every feasible $V$, with equality at $W$. Minimizing this minorant proves the upper bound. Evaluating $J$ at a minimizer of the minorant gives the lower bound. Thus an upper bound above tolerance can be inconclusive, while a positive lower bound produces an explicit improvement of the same admissible rank. Both bounds use products and inner products on $\mathbb S^d$. The spectral calculation only needs the lowest $r$ eigenvalues for the upper bound. In numerical work the accuracy of those eigenvalues must be accounted for when reporting a certificate.

\paragraph{A finite bound for logistic supervision.}
For a triplet with anchor $a$, positive $p$ and negative $n$, put $u=a-p$, $v=a-n$, $A_i=uu^\top-vv^\top$ and $b_i=1+\tr A_i$. For $m$ triplets and supervision weight $\eta>0$, we use
\begin{equation}
 h(W)=\frac{\eta}{m}\sum_{i=1}^m\log(1+\exp(b_i+\langle A_i,W\rangle)).
 \label{eq:logistic}
\end{equation}
For any probabilities $p_i\in[0,1]$, binary entropy $H(p)=-p\log p-(1-p)\log(1-p)$ gives a lower bound on $J_r^\star$:
\begin{align}
 B(p)&=s_r\!\left(C+\tfrac{\eta}{m}\textstyle\sum_i p_i A_i\right)
       +\tfrac{\eta}{m}\textstyle\sum_i[b_ip_i+H(p_i)]. \label{eq:dual}
\end{align}
This follows by replacing each softplus with a supporting line of slope $p_i$. At the logistic probabilities of $W$, denoted $p_W$, $J(W)-B(p_W)=\Delta_r(W)$. Maximizing $B(\theta p_W)$ over $0\le\theta\le1$ can tighten the bound and gives a finite choice at $\theta=0$ even when $K_W$ has a nonpositive eigenvalue. When $C=I$, the matrices along this segment share eigenvectors. A single eigendecomposition suffices for all values of $\theta$. Optimizing over all slopes need not close the gap over a nonconvex feasible set. Appendix~\ref{app:logistic-dual} gives a logistic example with strict slack at a global minimizer.

\section{When Does the Bound Vanish?}
Let $W=UU^\top$ with $U\in\mathbb R^{d\times r}$ and assume $h$ is twice continuously differentiable. A second-order stationary point is a factor at which the gradient vanishes and the factor Hessian is positive semidefinite. At a first-order stationary point of $U\mapsto J(UU^\top)$,
\begin{equation}
 G U=0,\qquad G=K_W-(I+W)^{-1}.\label{eq:kkt}
\end{equation}
In particular, the range of $W$ is invariant under $K_W$. A retained eigenvector of $W$ with mass $s>0$ is an eigenvector of $K_W$ with eigenvalue $1/(1+s)$. The unrestricted PSD optimality test additionally requires $G\succeq0$.

At a full-column-rank stationary point, let $s$ be the smallest positive eigenvalue of $W$, and let $k_b$ be the smallest eigenvalue of $K_W$ on the orthogonal complement of the range of $W$. The largest occupied eigenvalue of $K_W$ is $1/(1+s)$. The scalar minimizers in~\eqref{eq:oracle} then give the exact criterion
\[
 \Delta_r(W)=0 \quad\Longleftrightarrow\quad k_b\ge (1+s)^{-1}.
\]
Thus an omitted eigenvalue in $[(1+s)^{-1},1)$ can justify adding a column without justifying the replacement of an occupied one. The next result uses the curvature at a second-order point to force the required ordering.

\begin{theorem}[An exchange at the rank cutoff]\label{thm:exchange}
Let $U$ be a second-order stationary point and suppose $\nabla h$ is $L$-Lipschitz in Frobenius norm on $\mathbb S^d_+$. If $\rank U<r$, then $W$ is globally optimal over $\mathbb S^d_+$ and $\Delta_r(W)=0$. Otherwise let $s$ denote the smallest positive eigenvalue of $W$. If
\begin{equation}
 k_{r+1}(K_W)-k_r(K_W)>Ls,\label{eq:cut}
\end{equation}
then $K_W\succ0$, $\Delta_r(W)=0$, and $W$ is globally optimal over $\Kr$.
\end{theorem}

\begin{proof}
At rank deficiency, rotate the columns of $U$ to expose a zero column. Inserting $tz$ there changes $W$ by $t^2zz^\top$. Second-order stationarity gives $z^\top Gz\ge0$ for every $z$. Together with~\eqref{eq:kkt} this gives optimality for both $J$ and its convex supporting objective.

At full rank, the largest eigenvalue of $K_W$ on the retained subspace is $k_a=1/(1+s)$. If every omitted eigenvalue is at least $k_a$,~\eqref{eq:oracle} gives $W$ as a minimizer. Otherwise there are orthogonal unit vectors $u,v$ with $u$ retained at $k_a$ and $v$ omitted at the lowest omitted eigenvalue $k_b<k_a$. Rotate $u$ to $u\cos t+v\sin t$ by applying the ambient rotation to $U$. The eigenvalues of $W$ stay fixed. With $E=uv^\top+vu^\top$, the coefficient of $t^2$ in the change of $J$ is
\begin{equation}
 s(k_b-k_a)+\tfrac12s^2\nabla^2h(W)[E,E].\label{eq:dir}
\end{equation}
The second term is at most $Ls^2$. A misordered selection satisfies $k_a\ge k_{r+1}(K_W)$ and $k_b\le k_r(K_W)$ whenever this cutoff gap is positive. Equation~\eqref{eq:cut} would therefore give a negative second derivative, contradicting the hypothesis.
\end{proof}

Equation~\eqref{eq:dir} is often more informative than a uniform $L$. The pairwise curvature $\nabla^2h(W)[E,E]/2$ suffices in that exchange. The criterion also explains why the smallest retained mass enters the bound. It is the mass attached to the occupied mode that is easiest to replace. The support value~\eqref{eq:gap} can be evaluated even when the curvature test is inconclusive.

\paragraph{A uniform sufficient condition.}
Suppose $\|\nabla h(W)\|_2\le\delta$ for $W\succeq0$ and $\lambda_r(C)>\delta$, where $\lambda_i(C)$ are increasing. By Weyl's inequality and~\eqref{eq:kkt}, it suffices that
\[
 \lambda_{r+1}(C)-\lambda_r(C)>
 2\delta+L\big[1/(\lambda_r(C)-\delta)-1\big]_+ .
\]
Under this condition all second-order points are global at the specified rank. The uniform bound can be conservative. For example, an isotropic prior has zero eigengap in $C$, although supervision can induce a large gap in $K_W$ at a solution.

\paragraph{Computing the global minimizer.}
The spectral minimizer in~\eqref{eq:oracle} also defines an iteration.
Write $T(W)=V_{K_W}$ and $t(k)=(1/k-1)_+$.
\begin{theorem}[Convergence from any feasible start]\label{thm:iteration}
Suppose every $W\in\Kr$ satisfies $0<a\le k_1(K_W)$ and
$k_r(K_W)\le u<v\le k_{r+1}(K_W)$ for fixed $a\le u<v$.
Let $L$ be a Frobenius Lipschitz constant for $\nabla h$ on the PSD cone.
If $\rho=L\max\{a^{-2},t(u)/(v-u)\}<1$, then $W_{j+1}=T(W_j)$
converges from every $W_0\in\Kr$ to the unique global minimizer $W_*$.
In particular $\|W_j-W_*\|_F\le\rho^j\|W_0-W_*\|_F$.
\end{theorem}
The separation permits one scalar Lipschitz function to describe the
truncated update on both spectral bands. Its matrix extension obeys
the same Lipschitz bound in Frobenius norm. Thus $T$ is a contraction,
and its fixed point has $\Delta_r=0$. Appendix~\ref{app:iteration} proves uniqueness and bounds the distance to the solution using the last step. In a six-dimensional logistic example
with certified contraction factor below $0.718$, three distinct starts
all achieved a numerical gap below $10^{-7}$ after three updates.
Appendix~\ref{app:iteration-example} specifies this example. The metric-learning benchmarks use an isotropic prior and test the a posteriori bound directly.

\paragraph{Sharpness and the boundary.}
Take $d=2$, $r=1$, $C=\operatorname{diag}(c_1,c_2)$, $0<c_1<c_2<1$, and $h(W)=\kappa W_{12}^2$ with $\kappa\ge0$. Its Hessian has Frobenius operator norm $\kappa$. Put $s=1/c_2-1$. The point $W=s e_2e_2^\top$ is stationary. Rotating its direction toward $e_1$ gives the exact change
\begin{equation}
 s(c_1-c_2)\sin^2t+\kappa s^2\sin^2t\cos^2t .
 \label{eq:sharp}
\end{equation}
When $\kappa s<c_2-c_1$, the point has negative curvature. When $\kappa s>c_2-c_1$, the radial and angular curvatures are positive and the mixed derivative vanishes. This gives a strict local minimum of the rank-one factor objective. The global direction remains $e_1$, since it minimizes the unsupervised objective and has zero supervision cost. At equality the angular change is $-\kappa s^2\sin^4t$. The Hessian is positive semidefinite there, so replacing the strict comparison in~\eqref{eq:cut} with a weak one would give a false conclusion.

\paragraph{Other spectral losses.}
The exchange argument keeps the spectrum of $W$ fixed. It therefore also applies to $\tr\phi(W)+\langle C,W\rangle+h(W)$ for a twice continuously differentiable strictly convex scalar $\phi$, provided the scalar support minima are finite and attained. The active relation becomes $k_a=-\phi'(s)$, and the least retained mass still corresponds to the largest occupied $k_a$. Appendix~\ref{app:spectral} derives the certificate and relates it to squared-loss denoising. LogDet adds a closed form for its value and a natural reference metric.

\section{What a Fixed Agreement Penalty Guarantees}
For $X,Y\in\mathbb R^{d\times r}$ consider the explicitly symmetric extension
\begin{equation}
 M=\sym(XY^\top),\qquad
 F_\gamma(X,Y)=J(M)+\tfrac\gamma2\|X-Y\|_F^2 ,
 \label{eq:split}
\end{equation}
on $I+M\succ0$, with $\sym(Z)=(Z+Z^\top)/2$ and fixed $\gamma>0$. Specify $h$ on this domain as in~\eqref{eq:logistic}. A split output can always be converted into a feasible symmetric candidate, for example with $W=SS^\top$, $S=(X+Y)/2$. Its gap then concerns this particular $W$.

\begin{proposition}[Agreement and second-order transfer]\label{prop:split}
Assume $h$ is convex and twice differentiable on the split domain. If $\lambda_{\max}(C+\nabla h(M))\le B$ throughout that domain and $2\gamma\ge B-1$, all stationary points of~\eqref{eq:split} are diagonal. The infima agree. The split minimizers are exactly the pairs $(U,U)$ for which $UU^\top$ minimizes~\eqref{eq:model}. The second-order stationary points of~\eqref{eq:split} are exactly the diagonal pairs $(U,U)$ for which $U$ is second-order stationary for $U\mapsto J(UU^\top)$.
\end{proposition}

To see the source of the threshold, set $D=(X-Y)/2$, so $M=SS^\top-DD^\top$. If $G=\nabla J(M)$, stationarity gives $GS=0$ and $GD=2\gamma D$. Hence $S^\top D=0$. A nonzero eigenvalue $q$ of $DD^\top$ would yield a $-q$ eigenvalue of $M$ and a $2\gamma+1/(1-q)>2\gamma+1$ eigenvalue of $C+\nabla h(M)$, a contradiction. At equal factors, the Hessian in $(S,D)$ has zero mixed block. Its $S$ block is the symmetric factor Hessian. Its $D$ block is positive semidefinite because $G$ vanishes on the range of $S$ and $G\preceq(B-1)I$ on the orthogonal complement. Appendix~\ref{app:split} gives the block calculation. For the global comparison, the positive part $M_+$ has rank at most $r$, $\tr(M_-)\le\|D\|_F^2$, and removing a negative eigenvalue $-q$ saves $-\log(1-q)>q$ in the logarithmic term. Integration of the gradient of the remaining terms from $M_+$ to $M$ proves $F_\gamma(X,Y)\ge J(M_+)$.

Combining Proposition~\ref{prop:split} with Theorem~\ref{thm:exchange} gives a direct globality test for a split second-order point. Under the agreement threshold, write the point as $(U,U)$. If $\rank U<r$, or if $U$ has full column rank and its weakest retained mode satisfies~\eqref{eq:cut}, then $(U,U)$ globally minimizes $F_\gamma$ and $UU^\top$ globally minimizes $J$ at width $r$. The globality comparison is with every admissible pair of factors, including unequal pairs.

For logistic triplets a valid choice is $B=\lambda_{\max}(C+(\eta/m)\sum_i u_i u_i^\top)$. In the unsupervised case the cutoff $2\gamma\ge\lambda_{\max}(C)-1$ is sharp. For an eigenvalue $c>2\gamma+1$ with unit eigenvector $v$, setting $X=-Y=\sqrt{1-1/(c-2\gamma)}\,v$ in a single column gives an unequal stationary pair. The wrong direction in~\eqref{eq:sharp} is a strict local minimum of the split problem for \emph{every} $\gamma>0$ once supervision exceeds the curvature threshold. Thus increasing the penalty cannot resolve that subspace error.

\input{empirical}
\input{conclusion}
\renewcommand{\bibfont}{\small}
\setlength{\bibsep}{4pt plus 1pt minus 1pt}
\bibliographystyle{abbrvnat}
\bibliography{references}
\clearpage
\appendix
\section{Proofs and extensions}
\label{app:math}
\input{proofs}
\input{spectral_appendix}
\input{iteration_appendix}
\input{dual_appendix}
\section{Experimental methods and verification}
\label{app:experiments}
\input{synthetic_appendix}
\input{iteration_experiment}
\input{experiment_appendix}
\input{validated_appendix}
\input{trajectory_appendix}
\input{predictive_appendix}
\clearpage
\input{har_appendix}
\end{document}

%% file: related.tex
\section{Related Work}

\paragraph{Learning a metric with LogDet divergence.}
Information-theoretic metric learning fits a positive definite Mahalanobis matrix subject to pairwise distance constraints \citep{davis2007}. Triplet-based LogDet learning uses relative distances to supervise the metric \citep{mei2014}. Low-rank Bregman projections can preserve the column space of the initial kernel \citep{kulis2009}. Our metric has a full-rank identity baseline and a positive semidefinite update whose column space is learned. The distinction matters when the useful directions outnumber the available columns. \citet{bock2025} derive spectral truncations for low-rank approximations of a fixed matrix under LogDet divergence. In the supervised problem the matrix to be truncated changes with the candidate metric.

\paragraph{Optimality from factors.}
Theorem~2.1 of \citet{sahiner2024}, attributed there to \citet{mardani2013}, gives an a posteriori test for squared-loss denoising. At a stationary pair of Frobenius-regularized factors, $\|Y-UV^\top\|_2\le\beta$ certifies that $UV^\top$ minimizes the unrestricted objective $\tfrac12\|Z-Y\|_F^2+\beta\|Z\|_*$. Related trace-norm tests are studied by \citet{ciliberto2023}. Our comparison class is all PSD updates at the specified width, including widths at which the unrestricted optimum is unattainable. For a PSD factorization, rank deficiency at a second-order point provides a related certificate \citep{zhang2023cert}. \citet{ha2020} relate second-order stationarity of a factorization to fixed points of rank-constrained projected gradient.

\paragraph{Relation to asymmetric PSD factorization.}
The preprint by \citet{huPreprint} and the conference paper by \citet{hu2026} study agreement between the asymmetric and symmetric factorizations. Under their hypotheses, a sufficiently large fixed penalty forces the two factors to agree at critical points of the asymmetric problem. Agreement controls the difference between the two factors. Even with that difference equal to zero, a common factor may span a suboptimal subspace. The certificate studied here compares its objective value with the optimum over all admissible subspaces at the same rank budget. The curvature analysis identifies when exchanging an occupied direction for an unused one would lower the objective.

\paragraph{Bounds that retain the rank budget.}
The bound uses the affine minorant in the composite primal--dual gap of \citet[Eqs.~(16)--(17)]{yu2017}. Their suboptimality result assumes both components are convex. Here the rank restriction is nonconvex. Convexity of the supervised term gives a lower bound after minimization of its affine support together with the LogDet term over the prescribed rank budget. Convex-envelope constructions retain rank information for quadratic objectives \citep{grussler2018} and for classes of spectral epigraphs through matrix perspectives \citep{bertsimas2023}. We evaluate this bound by solving the rank-constrained LogDet subproblem and identify when a second-order point attains equality. Eigengaps also appear in self-consistent eigenproblems \citep{cai2018} and in spectral Frank--Wolfe analysis on convex spectrahedra \citep{ding2020}. In the present rank-constrained problem, the curvature of supervision enters through a rotation that exchanges an occupied direction and an omitted one. The two-dimensional example determines the constant in this comparison.

%% file: empirical.tex
\section{Experiments}
We ask whether the bound recognizes solutions missed by an unrestricted test, how informative it is when positive, and how often the curvature estimates explain a zero gap. Dataset-level counts refer to distinct optimization problems. Repeated solver outputs on one problem share its data and rank.

\begin{figure}[tbp]
\centering
\includegraphics[width=.97\linewidth]{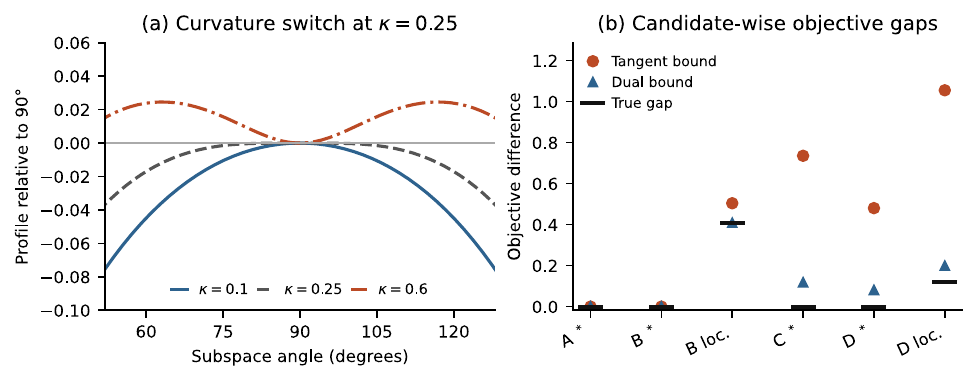}
\caption{Effect of supervision on a rank-one update. The left panel traces the objective around the suboptimal eigenvector at three supervision strengths. The right panel compares independently determined objective gaps with their upper bounds on four two-dimensional instances.}
\label{fig:landscape}
\end{figure}

\paragraph{Independent optima at rank one.}
Figure~\ref{fig:landscape} shows the curvature transition in~\eqref{eq:sharp} and a separate test with two noncommuting quadratic measurements. In the latter we minimize analytically over the radial variable and bound the error of an angular grid search independently of the certificate. The four instances have discretization error below $9.81\times10^{-9}$. On instance B the worse local minimum is $0.4091$ above the independently computed optimum and its tangent bound is $0.5044$. Instances C and D have globally best candidates to grid accuracy, yet their tangent bounds are $0.7359$ and $0.4805$. Appendix~\ref{app:synthetic} shows that optimizing both coefficients of the quadratic loss minorant still leaves positive gaps in these two cases. Absorbing a shared quadratic term into the spectral part before linearization yields bounds below $10^{-12}$ on the same candidates.

\input{experiments_main}

\paragraph{How much does the weakest mode matter?}
Of the 180 symmetric-factor outputs, 166 had numerical rank $r$, using a relative cutoff $10^{-8}$ for the smallest retained eigenvalue. The observed cutoff gap exceeded $L\|W\|_2$ on 40. Substituting the smallest retained mass, as in Theorem~\ref{thm:exchange}, raised the count to 84. Using the Hessian norm at the candidate gave 114, and the directional curvature from~\eqref{eq:dir} gave 154. Each of these 154 passed the objective-gap test. Among the 166 full-rank outputs, 160 passed it in total. These comparisons measure margins at numerical candidates. The objective-gap calculation itself requires no assumption on the Hessian of a factor iterate.

\paragraph{Cost of checking during training.}
Replaying all 60 Digits problems gave the same final objectives and evaluation counts as the original runs. Among the 54 problems that crossed tolerance, the median numbers of solver evaluations at first crossing were 22 for symmetric L-BFGS, 26.5 for split L-BFGS and 5 for projected gradient. The respective medians at normal termination were 37, 50 and 10.5. Auditing every accepted split step required a median of 24 additional matrix gradients at the feasible midpoint before first crossing. Symmetric and projected solvers could reuse their existing gradients. The six remaining problems terminated under the solvers' usual stopping rules without passing the test.

\paragraph{Held-out participants.}
We also used the official subject-disjoint split of the UCI smartphone activity recognition dataset \citep{ucihar}. We fit 64 principal components on the 7,352 training windows and evaluated on 2,947 windows from nine other participants. Three triplet draws, two negative-sampling rules, ranks 4 and 8, and weights 2 and 8 gave 24 training problems with 6,000 triplets each. Both the split and symmetric output passed the fixed-rank test on all 24. The unrestricted gradient sign test failed throughout. With hard negatives at $r=8$, $\eta=8$, mean held-out triplet ordering accuracy was 54.54\%, compared with 51.53\% for the unmodified Euclidean metric. The corresponding 5-NN accuracies were 87.99\% and 87.78\%. All three draws use the same held-out participants.

\paragraph{Prediction and training optimality.}
The accuracy differences in Table~\ref{tab:datasets} are descriptive. A separate comparison used supervised neighbourhood components analysis (NCA) \citep{goldberger2004} on the same splits. On WDBC, NCA gave 97.43\% and 97.08\% at widths two and four, compared with 96.61\% for both split models at $\eta=2$ with uniform negatives. On Digits, the full-rank Euclidean metric already gave 95.33\%. Appendix~\ref{app:predictive} reports the width-matched embeddings and an identity-plus-embedding comparison. Optimizing the sampled training objective and improving test accuracy are distinct questions.

%% file: experiments_main.tex
\paragraph{Metric learning on three data sets.}
We formed labeled triplets on Digits, Wine and Wisconsin Diagnostic Breast Cancer (WDBC) from scikit-learn \citep{sklearn,digits,wine,wdbc}. Five stratified splits, two negative-sampling rules, two ranks ($2,4$) and three supervision weights ($0.5,2,8$) gave 180 training problems. Each used $C=I$ and 2,400 triplets. The three iterative solvers shared the training triplets and rank. Symmetric-factor L-BFGS and projected gradient started from $S_0$ and $S_0S_0^\top$, respectively. For joint split-factor L-BFGS with fixed $\gamma$, the initial factor pair had midpoint $S_0$. A single projected step from the Euclidean metric provided a low-cost comparison. We audited every output independently at absolute objective-gap tolerance $10^{-7}$. The split output used $W=SS^\top$ with $S=(X+Y)/2$.

\begin{table}[t]
\centering
\setlength{\tabcolsep}{3pt}
\begin{tabular}{lrrrr}
\toprule
Data set & Uniform & Hard & Unrestricted & 5-NN change\\
\midrule
Digits & 25/30 & 29/30 & 0/60 & $+0.05$\\
Wine   & 30/30 & 30/30 & 38/60 & $+0.86$\\
WDBC   & 30/30 & 30/30 & 13/60 & $+0.92$\\
\bottomrule
\end{tabular}
\caption{Split-solver outputs satisfying the fixed-rank gap bound, grouped by sampling rule. The unrestricted column applies the same objective-gap test with rank budget $d$ across both rules. The last column is the mean change in held-out 5-NN accuracy from the Euclidean metric, in percentage points.}
\label{tab:datasets}
\end{table}

The double-precision fixed-rank test passed on the same 174 of 180 settings for all three iterative solvers (Table~\ref{tab:datasets}). Its count stayed at 174 for tolerances from $10^{-12}$ to $10^{-5}$. On the split outputs, the unrestricted bound met the $10^{-7}$ tolerance in 51 cases. We also reconstructed a thin factor from each stored split output. Taking the stored triplet differences and factor entries as exact, we enclosed $\Delta_r$ using 128-bit interval arithmetic (Appendix~\ref{app:validated}). For the same 174 problems this proves that the reconstructed update is within $7\times10^{-13}$ of its fixed-rank optimum. Split and symmetric final objective values differed by at most $6.9\times10^{-13}$. Projected gradient was fastest in these small dimensions, with median training time 2.38\,ms versus 10.57\,ms for symmetric and 13.24\,ms for split L-BFGS. The median endpoint audit took 0.19\,ms for the split method. These are single-thread timings excluding preprocessing. The one-step baseline certified none of the 180 settings. Its tangent matrix had a nonpositive eigenvalue in 40 cases, and the logistic ray bound remained finite in all 40.

\paragraph{A verified suboptimal rank-two update.}
The six settings with tangent gaps above tolerance all occurred on Digits at $r=2$, $\eta=8$. Their certificate values ranged from 0.0101 to 0.0524. We investigated these six settings with twelve additional symmetric-factor starts each. With seed 23 and hard negatives, the original two factor methods reached 9.302994 and projected gradient attained 9.302871 on the same objective. Nine restarts reached the lower value and three reached the higher value. The smallest computed Hessian eigenvalues orthogonal to factor rotations were 0.0159 at the lower representative and 0.0133 at the higher one. The original split factors differed by $1.24\times10^{-8}$ in Frobenius norm. On the archived triplets, an interval comparison of rank-two factors reconstructed from the split and projected solutions proved an improvement of at least $1.23377\times10^{-4}$ over the reconstructed split metric. The globality of the lower solution remains open. For each of the other five settings, all twelve additional starts agreed to seven decimal places.

%% file: conclusion.tex
\section{Discussion}
The example at the curvature boundary separates two possible failures in a split metric update. A weak agreement penalty can leave the factors unequal. Even when they are equal, supervision can stabilize an inferior subspace. The fixed-rank certificate tests the objective at the reported update and remains applicable to either factor-based or projected solvers. Which solver is cheapest depends on the dimension and the cost of the required gradients and eigenpairs.

A positive bound can reflect a better feasible update, slack from the affine support, or both. We observed both in problems with an independent reference or a lower-valued feasible competitor. One remaining question is how to retain some curvature of supervision in a tractable support problem at larger ranks. A second is how to certify large sparse metric updates with partial eigensolvers. The interval verification covers feasible factors reconstructed from all 180 split-solver outputs on the three smaller datasets. The activity recognition and solver-trajectory results use dense double-precision eigensolvers. Theorem~\ref{thm:iteration} gives a global iteration guarantee under uniform spectral bands. The benchmarked L-BFGS and projected-gradient runs are assessed by their endpoint bounds.

%% file: proofs.tex
\subsection{The fixed-rank lower bound}
\label{app:certificate}

We use the notation of Proposition~\ref{prop:gap}. Thus $1\le r<d$,
$\mathcal K_r=\{V\in\mathbb S^d:V\succeq0,\ 
\operatorname{rank}(V)\le r\}$, and
$J(V)=\langle C,V\rangle-\log\det(I+V)+h(V)$.
The inner product is $\langle A,B\rangle=\operatorname{tr}(A^\top B)$.
Take $h$ convex and differentiable on the domain in use. Write
$J_r^\star=\inf_{V\in\mathcal K_r}J(V)$.
For $W\in\mathcal K_r$, put $K=C+\nabla h(W)$ and
$q(V)=\langle K,V\rangle-\log\det(I+V)$.
The supporting inequality for $h$ gives, for every $V\in\mathcal K_r$,
\begin{equation}
 J(V)\ge J(W)+q(V)-q(W).
 \label{ap:eq:support}
\end{equation}
Hence $0\le J(W)-J_r^\star\le
\Delta_r(W):=q(W)-\inf_{V\in\mathcal K_r}q(V)$ whenever the right-hand
side is finite. No stationarity of $W$ enters this argument.

\paragraph{Evaluating the inner problem.}
Write $k_1\le\cdots\le k_d$ for the eigenvalues of $K$.
For fixed eigenvalues of $V$, the trace rearrangement inequality pairs
the largest eigenvalues of $V$ with the smallest eigenvalues of $K$
to minimize $\langle K,V\rangle$. Thus the inner optimization reduces
to at most $r$ scalar problems. If $k_1\le0$, taking
$V=t vv^\top$ along a unit $k_1$ eigenvector and sending $t$ to
infinity gives $\inf q=-\infty$, including when $k_1=0$.
For $k>0$, the unique minimizer of $kt-\log(1+t)$ on $t\ge0$ is
$t(k)=(k^{-1}-1)_+$. Its value is
\[
 b(k)=
 \begin{cases}
  1-k+\log k, & 0<k<1,\\
  0, & k\ge1.
 \end{cases}
\]
The function $b$ is nondecreasing, so
$\min_{V\in\mathcal K_r} q(V)=\sum_{i=1}^r b(k_i)$.
A minimizer is $V_0=\sum_{i=1}^r t(k_i)v_i v_i^\top$, where
the $v_i$ are corresponding orthonormal eigenvectors. Ties at the
cutoff allow any choice of the required number of directions within
the tied eigenspace. In numerical verification, a lower bound for
each of the $r$ smallest eigenvalues gives a lower bound for this
minimum if all of these bounds are positive. Uncorrected Rayleigh--Ritz
values can overestimate the smallest eigenvalues.

\paragraph{The loss omitted by linearization.}
For $D_h(V,W)=h(V)-h(W)-\langle\nabla h(W),V-W\rangle$, direct
subtraction gives
\[
 J(W)-J(V_0)=\Delta_r(W)-D_h(V_0,W).
\]
Thus $\max\{0,\Delta_r(W)-D_h(V_0,W)\}$ is a lower bound on the
actual objective gap. It is computable by evaluating $J(V_0)$.
If $\nabla h$ is $L$-Lipschitz in the Frobenius norm on the chord
$[W,V_0]$, $D_h(V_0,W)\le L\|V_0-W\|_F^2/2$.

\paragraph{A positive bound at a global optimum.}
Take $d\ge2$, $r=1$, $C=cI$ with $0<c<1$, and $h(V)=a\|V\|_F^2/2$
with $a>0$. Every matrix $W=p uu^\top$ with $\|u\|_2=1$ is globally
optimal when $p>0$ solves $(1+p)^{-1}=c+ap=:k_a$. Indeed the objective
at rank one reduces to $ct-\log(1+t)+at^2/2$ for $t\ge0$, with
unique minimizer $p$. The tangent matrix has eigenvalue $k_a$ on $u$
and $c$ on $u^\perp$. Therefore $\Delta_1(W)=b(k_a)-b(c)>0$.

\subsection{Curvature at fixed-rank stationary points}
\label{app:curvature}

Assume $h$ is twice continuously differentiable near the feasible
points considered. At $W=UU^\top$, put
$G=\nabla J(W)=K-(I+W)^{-1}$. Stationarity in $U$ is equivalent to
$GU=0$. Consequently $GW=WG=0$, and $K$ and $W$ have a common
orthonormal eigenbasis. Any positive eigenvalue $p_i$ of $W$ lies on
a direction $u_i$ with $K u_i=k_i u_i$, $k_i=1/(1+p_i)$. Eigenvalue
subscripts in the following exchange calculation identify directions;
we explicitly indicate ordered eigenvalues when using a cutoff.

\paragraph{Rank-deficient factors.}
For the first case of Theorem~\ref{thm:exchange}, suppose
$\operatorname{rank}U=\operatorname{rank}W<r$. There is a unit vector
$z\in\mathbb R^r$ with $Uz=0$. For every $v\in\mathbb R^d$, the factor
$U+t vz^\top$ has product $W+t^2vv^\top$. At a second-order stationary
point the quadratic coefficient $v^\top Gv$ is nonnegative. Hence
$G\succeq0$. Stationarity also gives $\langle G,W\rangle=0$, so
convexity certifies optimality over the PSD cone. On $\ker W$ we have
$K=G+I\succeq I$, while on the range of $W$ the active eigenvalues
satisfy $k_i<1$ and $p_i=t(k_i)$. The scalar characterization in
\eqref{eq:oracle} then gives $\Delta_r(W)=0$.

\paragraph{Exchanging occupied and unoccupied directions.}
Suppose $\operatorname{rank}W=r$. Rotate an occupied unit eigenvector
$u$ with eigenvalues $(p,k_a)$, $p>0$, toward an unoccupied unit
eigenvector $v$ of $K$ with eigenvalue $k_b$. All other eigenvectors
of $W$ are unchanged. The resulting $W(t)$ is a feasible curve and
can be obtained by an ambient orthogonal rotation of $U$. Since
$W(t)$ has the same spectrum as $W$, the log determinant is constant.
Write $H=uv^\top+vu^\top$. Then
\begin{align*}
 W'(0)&=pH,\\
 \langle K,W(t)-W\rangle&=p(k_b-k_a)\sin^2t.
\end{align*}
The expansion of $J(W(t))-J(W)$ has quadratic coefficient
\begin{equation}
 p(k_b-k_a)+p^2\nu_{uv},\qquad
 \nu_{uv}=\tfrac12\nabla^2h(W)[H,H].
 \label{ap:eq:exchange}
\end{equation}
A stationary factor has no second-order descent only if
$k_a-k_b\le p\nu_{uv}$ for each such pair, including pairs with
repeated eigenvalues. If the gradient of $h$ is $L$-Lipschitz, then
$0\le\nu_{uv}\le L$ since $\|H\|_F^2=2$.
The directional test can be evaluated with a Hessian-vector product.

Let $s=\lambda_r(W)>0$ denote the smallest retained eigenvalue, with
the eigenvalues of the PSD matrix $W$ ordered decreasingly.
The largest occupied eigenvalue of $K$ is $k_a=1/(1+s)$.
Write $k_b$ for the smallest eigenvalue of $K$ on $\ker W$.
Then $\Delta_r(W)=0$ precisely when $k_b\ge k_a$.
Indeed this is exactly the ordering of scalar modes required for
$W$ to minimize $q$. This assertion uses $\Delta_r=+\infty$ if
$K$ is not positive definite.

For comparison, denote the eigenvalues of $K$ in \emph{increasing}
order by $k_{(1)},\ldots,k_{(d)}$. If $k_b<k_a$, at least one of the
first $r$ ordered modes is unoccupied and at least one of the other
modes is occupied; ties can be resolved accordingly. Hence
$k_a\ge k_{(r+1)}$ and $k_b\le k_{(r)}$. Using the occupied direction
of mass $s$ in \eqref{ap:eq:exchange} gives the implication
\[
 k_{(r+1)}-k_{(r)}>Ls
 \quad\Longrightarrow\quad \Delta_r(W)=0
\]
at any full-rank second-order stationary point. Ties within an occupied
or unoccupied eigenspace cause no change to the argument.

\paragraph{A sufficient uniform condition.}
For this paragraph only, order the eigenvalues of $C$ increasingly as
$c_{(1)},\ldots,c_{(d)}$. Suppose $\|\nabla h(W)\|_2\le\delta$ at
all the PSD stationary points under consideration and the Hessian
upper bound $L$ holds there. Weyl's inequality gives
$k_{(r+1)}-k_{(r)}\ge c_{(r+1)}-c_{(r)}-2\delta$. The largest
occupied eigenvalue of a full-rank stationary point is at least
$k_{(r)}\ge c_{(r)}-\delta$. Thus if $c_{(r)}>\delta$,
$s\le[(c_{(r)}-\delta)^{-1}-1]_+$ for every such point. Combining
these estimates with the rank-deficient argument shows that every
second-order stationary point in the stated region is globally optimal
whenever
\begin{equation}
 c_{(r+1)}-c_{(r)}
 >2\delta+L\big[(c_{(r)}-\delta)^{-1}-1\big]_+ .
 \label{ap:eq:uniform}
\end{equation}
Here $[x]_+=\max(x,0)$. A common bound on $\|\nabla h\|_2$ is often
more conservative than a direct test of the final candidate.

\subsection{Sharpness at the curvature threshold}
\label{app:sharpness}

Take $d=2$, $r=1$, $C=\operatorname{diag}(c_1,c_2)$,
$0<c_1<c_2<1$, and $h(W)=\kappa W_{12}^2$ with $\kappa\ge0$.
In the space of symmetric matrices with the Frobenius norm, its
Hessian has operator norm $\kappa$ since
$2\kappa H_{12}^2\le\kappa\|H\|_F^2$, with equality for an
off-diagonal direction. The minimizer of the objective with $h=0$
is $W_1=(1/c_1-1)e_1e_1^\top$. Since $h$ is nonnegative and
vanishes at $W_1$, this is also a global minimizer of $J$.
The wrong stationary point is $W_2=p e_2e_2^\top$ with
$p=1/c_2-1>0$. Its objective exceeds the optimum by
$b(c_2)-b(c_1)>0$ for every $\kappa$.

In a neighborhood of $W_2$ parameterize rank-one factors by a
positive mass $z$ and a direction
$v(t)=e_2\cos t+e_1\sin t$, giving $W=zv(t)v(t)^\top$.
The objective at $(z,t)$ is
\[
 c_2z-\log(1+z)
 +z(c_1-c_2)\sin^2t+\kappa z^2\sin^2t\cos^2t .
\]
At $(p,0)$ the radial second derivative is $(1+p)^{-2}>0$ and the
mixed derivative vanishes. At fixed $z=p$, the change is exactly
\[
 p(c_1-c_2+\kappa p)\sin^2t-\kappa p^2\sin^4t.
\]
For $\kappa p<c_2-c_1$ the point has negative curvature. For
$\kappa p>c_2-c_1$ its Hessian is positive definite in these local
coordinates and it is a strict, nonglobal local minimum. At equality
it is a nonglobal second-order stationary point with quartic descent
along the displayed curve. Thus equality in the spectral-gap condition
does not suffice. The coordinates also apply locally to either factor
representing $W_2$, since the rank-one factor is nonzero.

\subsection{Transfer through the asymmetric factorization}
\label{app:split}

Suppose here that $h$ is convex and twice continuously differentiable
on $\{M\in\mathbb S^d:I+M\succ0\}$.
For $X,Y\in\mathbb R^{d\times r}$ define
$M=(XY^\top+YX^\top)/2$ and
$F_\gamma(X,Y)=J(M)+\gamma\|X-Y\|_F^2/2$, $\gamma>0$.
Let $S=(X+Y)/2$ and $D=(X-Y)/2$; this invertible change gives
\[
 M=SS^\top-DD^\top,\qquad
 F_\gamma=J(M)+2\gamma\|D\|_F^2 .
\]
At a stationary point, $G=\nabla J(M)$ is symmetric and
\[
 GS=0,\qquad GD=2\gamma D .
\]
The two eigenspaces are orthogonal, so $S^\top D=0$.
For any positive eigenvalue $q$ of $DD^\top$ and corresponding
unit eigenvector $v$, this gives $Mv=-qv$ with $0<q<1$.
Putting $K_M=C+\nabla h(M)$, stationarity requires
\[
 K_M v=\big(2\gamma+(1-q)^{-1}\big)v .
\]
Consequently, if $\lambda_{\max}(K_M)\le B$ on the split domain and
$2\gamma\ge B-1$, every stationary point has $D=0$.

\paragraph{Correspondence of second-order points.}
At a symmetric stationary factor $U$, the split point $(S,D)=(U,0)$
is stationary. Its Hessian in $(S,D)$ coordinates is block diagonal
because the mixed derivative vanishes at $D=0$. The $S$ block is the
Hessian of $U\mapsto J(UU^\top)$. In a difference direction $E$ the
quadratic coefficient is $\langle 2\gamma I-G,EE^\top\rangle$, with
$G=K_W-(I+W)^{-1}$ and $W=UU^\top$. On $\operatorname{range}(U)$
we have $G=0$. On its orthogonal complement, $(I+W)^{-1}=I$ and the
restriction of $G$ is bounded above by $(B-1)I$. Both subspaces
are invariant under $G$, hence
\[
 G\preceq\max\{0,B-1\}\,I\preceq 2\gamma I .
\]
The difference block is therefore positive semidefinite even at
$2\gamma=B-1$. Under the hypotheses of Proposition~\ref{prop:split},
the split second-order points are exactly the diagonal pairs $(U,U)$
for which $U$ is second-order stationary in the symmetric factor
problem. Since the added block has no negative eigenvalues and the change of
variables is invertible, the two Hessians have the same number of
negative eigenvalues, counted with multiplicity.

\paragraph{Comparison of the global objectives.}
Write $M=M_+-M_-$ for the positive and negative spectral parts, both
positive semidefinite. The representation $SS^\top-DD^\top$ gives
$\operatorname{rank}(M_+)\le r$ by the min--max principle and
$\operatorname{tr}(M_-)\le\|D\|_F^2$ by taking the trace on the
negative eigenspace. Joining $M_+$ to $M$ along $M_+-tM_-$ stays in
the domain. Integration along this chord shows
\begin{align*}
 &\langle C,M-M_+\rangle + h(M)-h(M_+)\\
 &\hspace{1.0em}\ge -B\operatorname{tr}(M_-).
\end{align*}
The log determinant contributes $-\log(1-q)>q$ for every negative
eigenvalue $-q$ of $M$. Hence $F_\gamma(X,Y)\ge J(M_+)$ when
$2\gamma\ge B-1$. The bound is strict for $M_-\ne0$, including
at equality in the penalty threshold. If $M_-=0$, the split penalty
is strictly positive unless $D=0$. Conversely every objective value
on $\mathcal K_r$ is attained by a diagonal pair $(U,U)$. This proves
equality of the infima and characterizes the minimizers when they
exist. In particular, for any feasible $W=UU^\top$ the bound on
$J(W)-J_r^\star$ also bounds the objective gap at $(U,U)$.

For the triplet loss in~\eqref{eq:logistic} with $\eta>0$,
write $a_i,b_i$ for the positive and negative pair differences.
The $i$th affine logit has gradient $a_i a_i^\top-b_i b_i^\top$.
The derivative of the scalar softplus lies between zero and one, so
\[
 B=\lambda_{\max}\!\left(C+\frac{\eta}{m}\sum_i a_i a_i^\top\right)
\]
is valid on the entire split domain and only uses training data.

\paragraph{The two-dimensional obstruction persists under splitting.}
At the stationary point $W_2$ of the preceding example,
$G=\operatorname{diag}(c_1-1,0)$. In $(S,D)$ coordinates the mixed
second derivative at $D=0$ vanishes and the $D$ quadratic
coefficient is $\langle 2\gamma I-G,DD^\top\rangle$, which is
positive for all nonzero $D$ and every $\gamma>0$. Thus $W_2$ lifts
to a strict nonglobal local minimum when $\kappa p>c_2-c_1$, and to
a nonglobal second-order stationary point with quartic descent at
equality. This local statement needs no uniform bound on $\nabla h$.

%% file: spectral_appendix.tex
\subsection{Extension to strictly convex spectral losses}
\label{app:spectral}
Write $g(W)=\operatorname{tr}\phi(W)+\langle C,W\rangle$, where
$\phi$ is scalar, twice continuously differentiable and strictly convex
on an open interval $\mathcal I$ containing zero. Restrict $W$ to matrices
with spectrum in $\mathcal I$. The matrix $C$ may be any symmetric matrix.
For $k\in\mathbb R$ let
\[
 b_\phi(k)=\inf_{t\in[0,\infty)\cap\mathcal I}
    \{kt+\phi(t)-\phi(0)\}.
\]
For the eigenvalues used in the support calculation, assume this infimum
is finite and attained. The minimizing $t$ is zero if $k+\phi'(0)\ge0$, and otherwise satisfies
$k+\phi'(t)=0$. Since $b_\phi(k)$ is nondecreasing in $k$, trace
rearrangement gives the support value
$d\phi(0)+\sum_{i=1}^r b_\phi(k_i)$ for increasingly ordered eigenvalues of
$K=C+\nabla h(W)$. The affine support inequality applies without change.
For a stationary factor $U$, $(K+\phi'(W))U=0$.
The map $t\mapsto-\phi'(t)$ is strictly decreasing, so the weakest retained
mode is paired with the largest occupied eigenvalue of $K$. The trace
of $\phi$ is constant along the rotation in Theorem~\ref{thm:exchange}.
The quadratic coefficient of the objective along that rotation
is therefore still given by~\eqref{ap:eq:exchange}. This gives the same threshold on the rank-cutoff gap.
At rank deficiency, a right-null vector of the factor again tests arbitrary
positive rank-one perturbations. Convexity then gives optimality over all
PSD matrices with spectrum in $\mathcal I$.
These conclusions concern globality for the specified objective and width.

\paragraph{The squared loss of the motivating residual test.}
For symmetric $Y$, minimizing $\|W-Y\|_F^2/2+\beta\operatorname{tr}W$ on the PSD cone is the PSD specialization of nuclear-norm denoising. It has $\phi(t)=t^2/2$, $C=\beta I-Y$, and $h=0$, up to a constant. For $Y=\operatorname{diag}(3,2)$, $\beta=1$ and $r=1$, the support oracle gives $W=\operatorname{diag}(2,0)$, with objective $4.5$ and fixed-rank gap zero. The residual satisfies $\|Y-W\|_2=2>\beta$. With no rank restriction, $\operatorname{diag}(2,1)$ attains $4$. This gives a direct comparison on the loss in Theorem~2.1 of \citet{sahiner2024}.

For the symmetric extension $M=SS^\top-DD^\top$ of~\eqref{eq:split},
assume that $\gamma>0$, all spectra stay in $\mathcal I$, and
$\lambda_{\max}(C+\nabla h(M))\le B$ throughout this domain. The
stationary equations are $GS=0$ and $GD=2\gamma D$ with
$G=C+\nabla h(M)+\phi'(M)$. A strictly negative eigenvalue $-q$ coming
from the difference factor gives an eigenvalue
$2\gamma-\phi'(-q)>2\gamma-\phi'(0)$ of $C+\nabla h(M)$.
Consequently $2\gamma\ge B+\phi'(0)$ forces equality of the factors.
Convexity gives $\phi(-q)-\phi(0)>-q\phi'(0)$ for $q>0$,
so the same condition also gives the global comparison with $M_+$.

%% file: iteration_appendix.tex
\subsection{A spectral iteration under uniform separation}
\label{app:iteration}

Assume the LogDet model in Appendix~\ref{app:certificate} and write
$K(W)=C+\nabla h(W)$. Suppose that every $W\in\mathcal K_r$ satisfies
\begin{align*}
 0<a&\le k_1(K(W)),\quad a\le u,\\
 k_r(K(W))&\le u<v\le k_{r+1}(K(W)),
\end{align*}
where eigenvalues are ordered increasingly. Let $t(k)=(1/k-1)_+$
for $k>0$, let $L$ be a Frobenius Lipschitz constant for $\nabla h$
on the PSD cone, and set
\[
 H=\max\{a^{-2},\,t(u)/(v-u)\},\qquad \rho=LH .
\]
If $\rho<1$, the iteration $W_{j+1}=T(W_j):=V_{K(W_j)}$ using
the minimizing spectral oracle of Appendix~\ref{app:certificate}, started from any $W_0\in\mathcal K_r$, converges
to the unique global minimizer $W_*$. Moreover
\begin{align*}
 \|W_j-W_*\|_F&\le \rho^j\|W_0-W_*\|_F,\\
 \|W_j-W_*\|_F&\le \|W_j-T(W_j)\|_F/(1-\rho).
\end{align*}
The bound $\Delta_r(W_j)$ remains available as an objective error bound
at each iterate; no monotone objective decrease is needed.

\paragraph{Proof.}
Define a scalar function $f$ equal to $t(a)$ below $a$, equal to
$t$ on $[a,u]$, affine between the values $t(u)$ and $0$ on $[u,v]$,
and zero above $v$. This is $H$-Lipschitz: $|t'(k)|\le a^{-2}$
wherever the derivative exists on $[a,u]$, and the middle linear segment
has absolute slope $t(u)/(v-u)$. By the uniform spectral separation,
$f(K(W))$ agrees with the unique minimizer of $q_{K(W)}$ on
$\mathcal K_r$. Indeed its nonzero eigenvalues are the scalar
minimizers on the first $r$ modes. No positive-mass tie is cut by
the width constraint. Ties among retained modes give the same matrix
because their optimal masses coincide.

For symmetric matrices $A,B$, let $(a_i,p_i)$ and $(b_j,v_j)$ be
orthonormal eigensystems. Expanding the Frobenius norm in these two
bases gives
\begin{align*}
 \|f(A)-f(B)\|_F^2
 &=\sum_{ij}|f(a_i)-f(b_j)|^2|p_i^\top v_j|^2\\
 &\le H^2\sum_{ij}|a_i-b_j|^2|p_i^\top v_j|^2\\
 &=H^2\|A-B\|_F^2 .
\end{align*}
It follows that $T(W)=f(K(W))$ is $\rho$-Lipschitz in the
Frobenius norm. It maps the closed set $\mathcal K_r$ into itself.
This set is complete, so Banach's theorem gives a unique fixed point
and the first rate bound. The triangle inequality applied to
$W-T(W)$ and $T(W)-T(W_*)$ gives the second. At the fixed point
$\Delta_r(W_*)=0$ proves global optimality. If $W'$ is another
global minimizer, the supporting inequality at $W_*$ implies
\[
 0\ge q_{K(W_*)}(W')-q_{K(W_*)}(W_*)\ge0 .
\]
The inner minimizer for $K(W_*)$ is unique, so $W'=W_*$.
This last step rules out optimizers at which a tangent bound
might otherwise have positive slack.

\paragraph{Data-dependent sufficient bounds.}
Suppose $\|\nabla h(W)\|_2\le\delta$ on $\mathcal K_r$.
Writing $c_1\le\cdots\le c_d$ for the eigenvalues of $C$, Weyl's
inequality permits $a=c_1-\delta$, $u=c_r+\delta$ and
$v=c_{r+1}-\delta$ whenever $a>0$ and $u<v$. Computing $L$ and $\delta$
from training data then gives a sufficient contraction test.
These uniform requirements can fail even when an individual tangent
bound certifies the solution. If the fixed-penalty condition
$2\gamma\ge B-1$ of Appendix~\ref{app:split} also holds on the
split domain, any factor $U$ of $W_*=UU^\top$ produces the global
split minimizer $(U,U)$. The iteration here acts directly on $W$.

%% file: dual_appendix.tex
\subsection{Finite logistic bounds and their limitations}
\label{app:logistic-dual}
Suppose that the supervision is
\[
h(V)=\frac{\eta}{m}\sum_{j=1}^{m}
\log\bigl(1+\exp(b_j+\langle A_j,V\rangle)\bigr),
\qquad \eta>0 .
\]
The additive offsets $b_j$ include the fixed identity contribution when the
learned distance matrix is $I+V$. Let $e(t)=-t\log t-(1-t)\log(1-t)$ for
$t\in[0,1]$, with $0\log 0=0$. For $z\in[0,1]^m$, set
\begin{align*}
K(z)&=C+\frac{\eta}{m}\sum_j z_j A_j,\\
a(z)&=\frac{\eta}{m}\sum_j (z_j b_j+e(z_j)),\\
B_r(z)&=a(z)+s_r(K(z)),
\end{align*}
where $s_r(K)$ is the minimum (or infimum) of
$\langle K,V\rangle-\log\det(I+V)$ over feasible PSD updates of rank at most $r$.
The scalar formula is $s_r(K)=\sum_{i=1}^r(1-k_i+\log k_i)
\mathbf{1}_{\{k_i<1\}}$ for positive eigenvalues $0<k_1\le\cdots\le k_d$;
$s_r(K)=-\infty$ if $k_1\le0$ and $r\ge1$.

\paragraph{Ray lower bound.}
For every feasible $W$ and every $z\in[0,1]^m$,
$B_r(z)\le J_r^\star\le J(W)$. Indeed
$\log(1+\exp t)=\max_{0\le u\le1}(ut+e(u))$. Applying this inequality
before minimizing over $V$ proves the assertion. Define
$p_j=1/(1+\exp[-b_j-\langle A_j,W\rangle])$.
At $z=p$, the affine minorant touches $h$ at $W$, hence
$J(W)-B_r(p)=\Delta_r(W)$. The function
$B_r$ is concave as the sum of a concave entropy and an infimum of affine
functions. If $C\succ0$ then $B_r(0)$ is finite. The bounded interval search
\[
\max_{0\le\theta\le1} B_r(\theta p)
\]
therefore gives a finite lower bound at least as large as $B_r(p)$, including
when $K(p)$ is indefinite. For $C=cI$, diagonalizing $\nabla h(W)$ once
gives the eigenvalues of $K(\theta p)$ for every $\theta$. For general $C$,
positive definiteness on the ray is equivalent to
$1+\theta\lambda_{\min}(C^{-1/2}\nabla h(W)C^{-1/2})>0$.
Numerical search need not attain the maximizing $\theta$ to give a bound:
evaluating any feasible point on the ray suffices in exact arithmetic.

\paragraph{An intrinsic gap.}
Even maximizing $B_r$ over the whole cube can leave a gap at a global
solution. Take $d=2$, $r=1$, $C=cI$ with $0<c<1$, and the four measurements
$A_j\in\{H_1,-H_1,H_2,-H_2\}$ with a common offset $b$, where
\[
H_1=\begin{pmatrix}1&0\\0&-1\end{pmatrix},\qquad
H_2=\begin{pmatrix}0&1\\1&0\end{pmatrix}.
\]
Each measurement is the difference of two rank-one PSD matrices. By strict
convexity of softplus, pairing opposite logits shows
$h(W)\ge\eta\log(1+\exp b)$, with equality exactly when $W$ is scalar.
Write $g(W)=\langle C,W\rangle-\log\det(I+W)$. Over PSD rank at most
one, $g$ has minimum $s_1(C)<0$, attained exactly on
$W=tuu^\top$ with $t=c^{-1}-1$ and $\|u\|=1$. For each $z$, evaluate the
affine minorant plus $g$ on both $t e_1e_1^\top$ and $t e_2e_2^\top$;
their average upper bounds its infimum. Its affine part at the average
$(t/2)I$ is at most $h((t/2)I)$. Hence
\[
\sup_z B_1(z)\le s_1(C)+\eta\log(1+\exp b).
\]
Equality follows by choosing all $z_j=(1+\exp[-b])^{-1}$.
The rank-one minimizers of $g$ are nonscalar, so none minimizes $h$.
Since $J$ is coercive on the closed feasible set, its attained minimum is
strictly larger than the displayed bound. A nonzero gap may thus persist at
a global solution of this deterministic two-dimensional problem.

%% file: synthetic_appendix.tex
\subsection{Independent two-dimensional reference values}
\label{app:synthetic}
We used four two-dimensional rank-one instances to distinguish a positive
certificate at a suboptimal local minimum from a positive certificate at a
global minimum. The two measurements do not commute. With
$C=\operatorname{diag}(1/4,1/2)$ and $W\succeq0$, they are
\[
b_1(W)=W_{12},\qquad b_2(W)=(W_{11}-W_{22})/2 .
\]
The supervised term is $h(W)=\sum_{j=1}^2 k_j(b_j(W)-d_j)^2$.
The columns give the two weights and two targets for each of the four deterministic instances.
\[
\begin{array}{c|rrrr}
  & k_1 & k_2 & d_1 & d_2 \\
  \hline
  \mathrm{A} & 0.10 & 0.015 & 0.08 & 0.05 \\
  \mathrm{B} & 0.85 & 0.02  & 0.08 & 0.05 \\
  \mathrm{C} & 0.80 & 0.80  & 0.12 & -0.07 \\
  \mathrm{D} & 1.60 & 0.45  & 0.10 & -0.10
\end{array}
\]

\paragraph{Independent reference interval.}
Set $W=t v_\theta v_\theta^\top$, $v_\theta=(\cos\theta,\sin\theta)$,
$t\geq0$, $0\leq\theta<\pi$. Write
$a=v_\theta^\top C v_\theta$ and
$b=(\sin(2\theta),\cos(2\theta))/2$. The objective at fixed angle is
\[
F(t,\theta)=at-\log(1+t)+\sum_j k_j(tb_j-d_j)^2 .
\]
It is strictly convex in $t$. Its minimum is attained at zero if
$a-2\sum_j k_j b_j d_j\geq1$, and otherwise at the unique positive
root of a quadratic obtained from $(1+t)\partial_tF=0$.
All optimal radii are less than $T=4$: at $t=T$ the radial derivative
is at least $1/4-1/(1+T)-\sum_j k_jd_j^2/(2T)>0$.
For fixed $t\leq T$, a uniform upper bound on the angular second
derivative is
\[
H=T/2+2\sum_j k_j\{T^2+2T(T/2+|d_j|)\}.
\]
Indeed $|a''|\leq 1/2$, $|b_j'|\leq1$, and $|b_j''|\leq2$.
At a global minimizing pair the angular first derivative vanishes.
Rounding its angle to the nearest point of the uniform $N=2^{17}$ grid
thus increases the objective by at most $H\pi^2/(8N^2)$.
If $v_N$ is the smallest radially minimized grid value, the global
minimum lies in $[v_N-H\pi^2/(8N^2),v_N]$.
Refining each detected minimum can lower the upper endpoint. Across
the four cases the mesh error is below $9.81\times10^{-9}$. This bounds
discretization error; computations use floating-point arithmetic.

\begin{table}[tb]
\centering
\small
\begin{tabular}{lrrr}
\hline
Candidate & Reference gap & Tangent bound & Dual bound\\
\hline
A, best & 0.0000 & 0.0000 & 0.0000\\
B, best & 0.0000 & 0.0000 & 0.0000\\
B, other & 0.4091 & 0.5044 & 0.4091\\
C, best & 0.0000 & 0.7359 & 0.1186\\
D, best & 0.0000 & 0.4805 & 0.0806\\
D, other & 0.1187 & 1.0554 & 0.1993\\
\hline
\end{tabular}
\caption{Fixed-rank objective gaps on the noncommuting examples.
The independent reference intervals have widths below $10^{-8}$.
The dual bound uses two freely chosen linearization coefficients.}
\label{tab:synthetic-gaps}
\end{table}

\paragraph{Why optimizing the linearization can leave a gap.}
Let $B_j$ represent $b_j$ by the trace inner product. For any
$s\in\mathbb R^2$, the rank-one spectral minimum $s_1$ from
Appendix~\ref{app:certificate} gives the lower bound
\[
s_1\!\left(C+\sum_j s_j B_j\right)
-\sum_j\left(s_jd_j+\frac{s_j^2}{4k_j}\right).
\]
Taking $s_j=2k_j(b_j(W)-d_j)$ gives the tangent bound in the table.
Every finite evaluation in the two-dimensional dual search is a valid
lower bound. For C and D, even the best choice of $s$ leaves the displayed
duality gap. At $s=(0,1/4)$ the effective matrix is $(3/8)I$,
so every direction minimizes the inner problem at radius $5/3$.
The convex hull of their measurement pairs is the disk of radius $5/6$.
In both cases $(d_j+s_j/(2k_j))_{j=1}^2$ lies inside that disk;
the concave dual objective therefore has zero in its superdifferential.

One can verify that the gap depends on the chosen decomposition of the
objective. On the rank-one feasible set,
$b_1(W)^2+b_2(W)^2=\operatorname{tr}(W^2)/4$. For $m=\min_j k_j$,
absorb $m\operatorname{tr}(W^2)/4$ into the spectral term before
linearizing the remaining convex term. This alternate bound is below
$10^{-12}$ at the best candidates in C and D. The identity uses these
particular measurements and dimension; the real-data experiments use the
original loss decomposition.

%% file: iteration_experiment.tex
\subsection{Contractive spectral iteration}
\label{app:iteration-example}
We tested the sufficient contraction condition on a six-dimensional
rank-two problem with
$C=\operatorname{diag}(.55,.62,1.30,1.50,1.65,1.90)$.
Let $A_i=v_iv_i^\top$, where $v_1=(e_1+e_3)/\sqrt2$ and
$v_2=(e_1+e_4)/\sqrt2$, and set
$h(W)=(.36/2)\sum_{i=1}^2\log(1+\exp(b_i+\langle A_i,W\rangle))$
with $b=(-.35,.45)$. The measurements do not commute.
The sigmoid lies in $[0,1]$ and its derivative is at most $1/4$.
Thus $\delta=.36\|(A_1+A_2)/2\|_2=.27$ and
$L=(.36/8)\lambda_{\max}([\langle A_i,A_j\rangle]_{ij})=.05625$
are valid uniform bounds. The corresponding band parameters are
$(a,u,v)=(.28,.89,1.03)$ and the contraction factor is at most
$.7175$. We initialized at zero, at $3e_3e_3^\top+2e_4e_4^\top$,
and at the solution for $h=0$. Their initial tangent gaps were
$.1861$, $4.9303$, and $.0215$, respectively. Each first crossed
$10^{-7}$ after three updates (four gradient and eigensystem evaluations,
including the final audit). The objective agreed to twelve decimal
places across the three starts. This constructed example tests the
stated sufficient condition.

%% file: experiment_appendix.tex
\subsection{Metric-learning experimental protocol}
\label{app:experiment}
\paragraph{Data and partitions.}
We used the copies of Digits (1,797 observations, 64 features, 10 classes), Wine (178, 13, 3), and Wisconsin Diagnostic Breast Cancer, or WDBC (569, 30, 2), shipped with scikit-learn. Digits is the 1,797-observation copy of Optical Recognition of Handwritten Digits. These small image and tabular benchmarks cover both multiclass and binary triplets, and permit independent checks using the full spectrum. Each data set used five stratified 70/30 train/test splits with seeds $11,23,37,53,71$. This gives train/test sample counts of $1257/540$, $124/54$ and $398/171$, respectively. We standardized each coordinate using the training-set mean and standard deviation, fit PCA on the training set alone, and retained $d=\min(16,d_{\mathrm{raw}})$ coordinates. Each observation in this representation was divided by its Euclidean norm (a zero row would remain zero). No labels from the test split entered preprocessing, triplet construction, training or model choice.

\paragraph{Triplet construction and objective.}
For each training split we sampled $m=2400$ anchors uniformly with replacement. For each anchor we sampled a distinct same-class positive and eight other-class negative candidates uniformly with replacement. The uniform protocol used the first negative candidate. The hard protocol used the nearest of the eight in the initial Euclidean metric. Both protocols shared the anchors, positives and candidate lists. All solvers, ranks $r\in\{2,4\}$, and weights $\eta\in\{0.5,2,8\}$ shared the triplets from a given protocol and split. Let $a_i=x_i-x_i^+$, $b_i=x_i-x_i^-$, and $A_i=a_i a_i^\top-b_i b_i^\top$. We minimized
\[
 \begin{aligned}
 &\operatorname{tr} W-\log\det(I+W)\\
 &\quad +\frac{\eta}{m}\sum_{i=1}^m
 \log(1+\exp(1+\langle I+W,A_i\rangle)).
 \end{aligned}
\]
All six rank/weight combinations contribute fully to the results. Signed matrices $A_i$ were vectorized with off-diagonal entries scaled by $\sqrt{2}$, preserving Frobenius inner products. With $D$ the matrix of these row vectors, we used $L_h=\eta\lambda_{\max}(D^\top D)/(4m)$ as a global upper bound on the Lipschitz constant of the logistic gradient.

\paragraph{Solvers.}
Symmetric L-BFGS optimized $W=UU^\top$. Split L-BFGS optimized $M=(XY^\top+YX^\top)/2$ jointly over both factors, with penalty $\gamma\|X-Y\|_F^2/2$. Its constant value in each run was
\[
\gamma=0.05+\frac{\eta}{2}\lambda_{\max}\!\left(m^{-1}\sum_i a_i a_i^\top\right).
\]
Indeed $\nabla h(M)\preceq(\eta/m)\sum_i a_i a_i^\top$ since every sigmoid weight belongs to $[0,1]$. This gives a uniform upper bound needed to eliminate unequal stationary factors in the split model. The split factors started at $(S_0+D_0,S_0-D_0)$ with independent centered normal entries of standard deviations $0.12$ and $0.06$. The symmetric solver started at the same $S_0$ and projected gradient started at $S_0S_0^\top$. Rank and split seed fixed each initialization before solver comparisons. The two L-BFGS runs used SciPy \citep{scipy}, a 500-evaluation target, maximum 500 iterations, 50 line-search trials, relative function tolerance $10^{-15}$ and factor gradient tolerance $10^{-9}$. Split trials for which $I+M$ failed Cholesky factorization received a large finite rejection value. All returned endpoints were feasible. We evaluated the split output at $SS^\top$, $S=(X+Y)/2$.

Projected gradient used truncation to the $r$ largest positive eigenvalues after a gradient step. The starting step was $(1+L_h)^{-1}$. Subsequent steps used the ratio $\|\Delta W\|_F^2/\langle\Delta W,\Delta G\rangle$, clipped to $[10^{-4},50]$ with a positive denominator floor of $10^{-20}$. Trials were halved until the objective decreased by at least $10^{-4}\|\Delta W\|_F^2/\alpha$ to floating-point precision at step $\alpha$. We stopped at projected residual $\|\Delta W\|_F/\alpha\le 10^{-8}$ or 500 objective and gradient evaluations. The one-step method made one projected update directly from $W=0$ with step $(1+L_h)^{-1}$. This method measures how much a single gradient-informed spectral update achieves before fitting.

\begin{table}[t]
\centering
\setlength{\tabcolsep}{3pt}
\begin{tabular}{lrrrrr}
\toprule
Solver & Rank $r$ & Rank $d$ & Evals. & Fit (ms) & Audit (ms)\\
\midrule
Symmetric & 174 & 51 & 60 & 10.57 & 0.19 \\
Split & 174 & 51 & 73 & 13.24 & 0.19 \\
Projected & 174 & 51 & 12 & 2.38 & 0.16 \\
One step & 0 & 0 & 1 & 0.22 & 0.18 \\
\bottomrule
\end{tabular}
\caption{Number of training problems with objective-gap bound at most $10^{-7}$ out of 180 per solver. Work and times are medians. Audit time includes a fresh objective and gradient at the PSD output and an eigendecomposition of the supporting matrix.}
\label{tab:solver-results}
\end{table}

\paragraph{Coverage and timing.}
The rank $d$ comparison uses exactly the same convex support and the same absolute tolerance as the rank $r$ test. Hence it tests objective gaps in the same units. For comparison, the distinct necessary sign condition $\lambda_{\min}(\nabla J(W))\ge-10^{-7}$ holds at 28 of the 180 split endpoints. Floating-point differences at the scale of roundoff are included in the reported numerical certifications. The solver comparisons use dense double-precision symmetric eigensolvers. Appendix~\ref{app:validated} gives a separate interval verification for all 180 split outputs on these datasets. Accuracy is the fraction of correctly classified held-out observations under exact 5-nearest-neighbor prediction, with distances computed using $I+W$. Each split is reused across optimization conditions and these accuracies are descriptive, without a claim of independent replications across conditions. The mean Euclidean and split accuracies in percent were $95.33/95.38$ for Digits, $96.67/97.53$ for Wine, and $95.79/96.71$ for WDBC.

Timing used a single BLAS thread on an AMD EPYC 9V45 CPU under Linux (9.73 GiB of visible memory). The environment was Python 3.12.14, NumPy 2.3.5, SciPy 1.17.0, scikit-learn 1.8.0 and OpenBLAS 0.3.30. Optimization timers exclude data preparation, model construction, and test prediction. The endpoint audit was timed independently. In these low dimensions, median times and absolute tolerances provide more useful comparisons than asymptotic speed claims.

\paragraph{Restarts and local curvature.}
For every split endpoint whose fixed-rank bound exceeded $10^{-7}$, we ran twelve additional symmetric fits initialized independently, using seeds $20000$ through $20011$. This diagnostic followed the single-start comparison and contributed no runs to its coverage counts. For $W=UU^\top$ write $E=UT^\top+TU^\top$. The symmetric factor Hessian applied to $T$ is
\[
 2\left(\nabla J(W)T+\nabla^2J(W)[E]U\right).
\]
We formed this small Hessian in a coordinate basis and compressed it to the Frobenius-orthogonal complement of $\{U\Omega:\Omega^\top=-\Omega\}$, the infinitesimal rotation directions. On the split with seed 23 and hard negatives, the two objective values differed by $1.23\times10^{-4}$. Both sets of numerical representatives had positive curvature on this complement (at least $0.0132$), and the maximum absolute gradient row sum over the twelve restarts was below $6.5\times10^{-8}$. Positivity at a numerical representative is a diagnostic. The independently attainable lower objective is the evidence of suboptimality of the higher endpoint. The remaining five unresolved configurations gave no better competitor among these starts.

%% file: validated_appendix.tex
\subsection{Interval validation of the split-solver outputs}
\label{app:validated}
We checked all 180 split-solver outputs from the three-dataset benchmark
using real ball arithmetic \citep{johansson2017arb}. Each verification problem is specified by
the binary64 entries of the archived positive and negative input
differences and a thin factor $U$. We obtained $U$ by taking the leading
$r$ eigenpairs of the saved midpoint product, truncating their eigenvalues
at zero, and scaling each eigenvector by the square root of its truncated
eigenvalue. We stored the resulting factor in binary64. The exact real product $UU^\top$ of these dyadic
entries is feasible by construction. The resulting optimization problem treats the stored differences as exact input data. This convention fixes the problem after preprocessing and the feasible point after training.

At 128 bits of precision, we enclosed each squared distance, logistic
probability, and entry of $K_W$. We used the Rump algorithm \citep[Sec.~13.4]{rump2010}, implemented in
\texttt{python-flint} 0.9.0, to compute eigenvalue balls and checked that
the real enclosures were pairwise disjoint and positive. For each of the
smallest $r$ eigenvalues we also checked on which side of $1$ it lay.
Finally,
$\det(I+UU^\top)=\det(I+U^\top U)$ gave the log determinant from an
$r$-dimensional matrix. Each numerical quantity entering
$\Delta_r(W)$ was computed with ball arithmetic. The CSV records both
endpoints, rounded outwards.

\begin{center}
\begin{tabular}{lrr}
Dataset & Below $10^{-7}$ & Largest passing bound \\
\hline
Digits & $54/60$ & $1.13\,10^{-14}$ \\
Wine & $60/60$ & $6.95\,10^{-13}$ \\
Breast Cancer & $60/60$ & $1.20\,10^{-13}$ \\
\end{tabular}
\end{center}
On all 174 passing instances, the verified upper bound on the true
fixed-rank suboptimality was below $7\times10^{-13}$. The remaining
six intervals for $\Delta_r$ lay within $[0.0101,0.0524]$, all on
Digits at rank two and $\eta=8$. Even at a global minimizer the tangent
certificate may be positive. Establishing suboptimality therefore requires
a separate lower bound on objective error, as in the example below. Every
interval lay on one side of $10^{-7}$, and all classifications agreed with
the double-precision audit. The validation
script takes the archived dyadic inputs directly.

\paragraph{A two-sided bound on actual suboptimality.}
For seed 23 and hard negatives, we reconstructed a second rank-two
factor from the projected-gradient output. Evaluating both feasible
factor products on the same exact triplets enclosed
$J(W_{\rm split})-J(W_{\rm proj})$ between
$0.00012337710718387647$ and $0.00012337710718387653$.
Since $J_2^\star\le J(W_{\rm proj})$, this provides a lower bound
on the objective error of the reconstructed split metric. Its independently enclosed
tangent gap supplies an upper bound. Rounded outwards, they give
\[
 0.0001233771\;\le\; J(W_{\rm split})-J_2^\star
 \;\le\;0.010116780.
\]
The two enclosures use the same split factor. This comparison leaves
the projected candidate's global optimality open.

%% file: trajectory_appendix.tex
\subsection{Certification along optimization trajectories}
\label{app:trajectories}
We replayed all 60 Digits settings with each of the three iterative
solvers, using identical triplets, starting points, line searches and
stopping rules. At the initial point and after every accepted step we
evaluated the tangent gap in double precision at tolerance $10^{-7}$. Rejected line-search
trials count as objective/gradient evaluations and are excluded from
the accepted-step count. The trace did not affect stopping. All 180
final objective values and evaluation counts matched the original
runs. The same 54 settings crossed the threshold for all three
solvers. The other six stopped by their usual rules without crossing
it. Each used fewer than 500 evaluations.

Among the common 54, the median solver evaluation counts at first crossing
were 22 (symmetric L-BFGS), 26.5 (split L-BFGS), and 5 (projected gradient).
The corresponding median evaluation counts at solver termination were
37, 50, and 10.5. The symmetric and projected solvers reused the matrix
gradient at the accepted PSD point. For split L-BFGS, evaluating the
feasible midpoint product required a median of 24 extra matrix gradient
evaluations by first crossing, in addition to the reported solver calls.
In this single-threaded replay, the summed audit time divided by summed
solver-plus-audit time over all 60 fits was 10.9\%, 42.3\%, and 7.4\%,
respectively. The timing measures this implementation and audit
frequency. The per-step and per-run traces accompany the code.

%% file: predictive_appendix.tex
\subsection{Predictive comparisons at matched update width}
\label{app:predictive}
We also evaluated PCA and Neighborhood Components Analysis (NCA) \citep{goldberger2004} on the five stratified splits of each small dataset. All methods used the same standardized, PCA-preprocessed, row-normalized observations and the same held-out 5-nearest-neighbor vote. The additional PCA baseline projects these observations onto their top $r$ training principal components. NCA learns a linear embedding $A$ with $r$ rows using all training labels and its stochastic nearest-neighbor objective. We initialized $A$ with the same PCA directions and set a budget of 50 L-BFGS iterations with tolerance $10^{-5}$. Two of the 30 fits reached the iteration limit. The rank-update model used the uniform triplets with the middle supervision weight $\eta=2$ in every comparison. The test set was used only for scoring.

\begin{table}[ht]
\centering
\caption{Mean held-out 5-NN accuracy (\%) over five splits. Base uses the identity metric, NCA uses $A^\top A$, NCA+ uses $I+A^\top A$ from the same fit, and Split uses $I+SS^\top$.}
\label{tab:predictive-extra}
\small
\begin{tabular}{@{}lr rrrrr@{}}
\toprule
Data & $r$ & Base & PCA & NCA & NCA+ & Split\\
\midrule
Digits & 2 & 95.33 & 58.30 & 62.85 & 62.89 & 95.30\\
       & 4 & 95.33 & 81.04 & 86.81 & 86.85 & 95.30\\
Wine   & 2 & 96.67 & 97.41 & 96.67 & 95.93 & 97.04\\
       & 4 & 96.67 & 93.70 & 96.67 & 96.30 & 97.04\\
WDBC   & 2 & 95.79 & 93.68 & 97.43 & 97.43 & 96.61\\
       & 4 & 95.79 & 94.97 & 97.08 & 97.08 & 96.61\\
\bottomrule
\end{tabular}
\end{table}

Training uses different supervision in the two approaches. NCA evaluates all training neighbors, whereas Split fits a sampled triplet loss with LogDet regularization. On Digits the Euclidean baseline already attains 95.33\%, and a rank-two embedding discards much of that discriminative information. On WDBC, NCA attains the highest accuracy at both widths. These comparisons describe prediction on the held-out splits. The objective-gap bounds refer to the regularized training objective.

At $r=4$ and $\eta=2$ with uniform negatives, we subtracted the Euclidean accuracy from the split accuracy separately on each of the five test splits. The mean paired changes (sample standard deviations), in percentage points, were $-0.04$ ($0.15$) for Digits, $+0.37$ ($0.83$) for Wine, and $+0.82$ ($0.67$) for WDBC. The split metric improved on one Wine split and tied on four. It improved on four WDBC splits and tied on one. These spreads describe variation across the five overlapping stratified splits.

%% file: har_appendix.tex
\subsection{Subject-disjoint smartphone activity benchmark}
\label{app:har}
The UCI Human Activity Recognition Using Smartphones data provide
561 sensor-derived features and six activity labels per window
\citep{ucihar}. The official split assigns 7,352 training windows from 21
people and 2,947 test windows from nine other people. Feature standardization and 64-dimensional PCA used only
training windows. We normalized each resulting row. The three seeds
11, 23 and 37 vary the triplets and initialization on this single subject partition. Per seed, we sampled 6,000
training triplets and separately sampled 6,000 held-out triplets. In both
partitions, positives were distinct windows of the anchor's class. The
uniform negative was the first of eight uniformly sampled other-class
candidates, and the hard negative was the nearest of the same eight
under the normalized Euclidean metric. Held-out triplet labels entered
only the evaluation. We used the same training loss, penalty rule,
single-thread budget, and symmetric and split initializations as in the
smaller experiments, here with ranks 4 and 8 and weights 2 and 8.
Factor gradients used matrix--factor products.

\begin{table}[htbp]
\centering
\small
\begin{tabular}{lrrrr}
\hline
Negatives & $r$ & $\eta$ & Ordering (\%) & 5-NN (\%)\\
\hline
Uniform & 4 & 2 & 88.18 & 87.74\\
Uniform & 4 & 8 & 88.66 & 87.71\\
Uniform & 8 & 2 & 88.19 & 87.74\\
Uniform & 8 & 8 & 88.67 & 87.71\\
Hard    & 4 & 2 & 52.17 & 87.74\\
Hard    & 4 & 8 & 54.53 & 87.93\\
Hard    & 8 & 2 & 52.17 & 87.80\\
Hard    & 8 & 8 & 54.54 & 87.99\\
\hline
\end{tabular}
\caption{UCI HAR held-out measurements for split fits, averaged over
the three seeds. Ordering is the fraction of test triplets with the
positive strictly closer than the negative. The Euclidean baseline
achieves 88.14\% uniform ordering, 51.53\% hard ordering, and 87.78\%
5-NN accuracy.}
\label{tab:har-results}
\end{table}

Evaluated in double precision, all 24 split and 24 symmetric fits gave tangent and ray gaps below
$10^{-7}$. The largest split ray gap was $1.34\times10^{-12}$. Every
unrestricted gradient sign test failed at tolerance $10^{-7}$.
The maximum absolute difference between paired symmetric and split
objectives was $1.18\times10^{-12}$. The hard-negative model improves
held-out ordering at the larger weight, while 5-NN accuracies across
these settings remain close to the Euclidean value.